\documentclass[10pt,journal,compsoc]{IEEEtran}
\ifCLASSOPTIONcompsoc
  \usepackage[nocompress]{cite}
\else
  \usepackage{cite}
\fi

\usepackage{booktabs} 
\usepackage{times}
\usepackage{graphicx}
\usepackage{graphics}
\usepackage{comment}
\usepackage{amsmath}
\usepackage{mathtools}
\usepackage{amsfonts}
\usepackage{multirow}
\usepackage{algorithm}
\usepackage{algorithmic}

\usepackage{epsfig}

\usepackage{amssymb}
\usepackage{makecell}
\usepackage{balance}
\usepackage{color}

\newcommand{\bx}{\textbf{x}}
\newcommand{\by}{\textbf{y}}

\newcommand{\bM}{\textbf{M}}

\newcommand{\bG}{\textbf{G}}

\ifCLASSINFOpdf
\else
\fi
\begin{document}
%
\title{Visual Font Pairing}
%
%
%
%

\author{Shuhui~Jiang,
        Zhaowen~Wang,~\IEEEmembership{Member,~IEEE},
        Aaron~Hertzmann,~\IEEEmembership{Senior Member,~IEEE},
        Hailin~Jin, ~\IEEEmembership{Member,~IEEE},
        and~Yun~Fu,~\IEEEmembership{Senior Member,~IEEE}
\IEEEcompsocitemizethanks{\IEEEcompsocthanksitem S. Jiang is with the Department of Electrical and Computer Engineering, Northeastern University, Boston, MA 02115 USA (e-mail: shjiang@ece.neu.edu). Y. Fu is with the Department of Electrical and Computer Engineering,College of Engineering, and College of Computer and Information Science, Northeastern University, Boston, MA 02115 USA (e-mail: yunfu@ece.neu.edu). Z. Wang, A. Hertzmann and H. Jin are with Adobe Systems Inc. (e-mail: \{zhawang, hertzman, hljin\}@adobe.com). }
\thanks{Manuscript received April 19, 2005; revised August 26, 2015.}}

%
%

\markboth{Journal of \LaTeX\ Class Files,~Vol.~14, No.~8, August~2015}%
{Shell \MakeLowercase{\textit{et al.}}: Bare Demo of IEEEtran.cls for Computer Society Journals}
%



\IEEEtitleabstractindextext{%
\begin{abstract}
This paper introduces the problem of \textbf{automatic font pairing}. Font pairing is an important design task that is difficult for novices. Given a font selection for one part of a document (e.g., header), our goal is to recommend a font to be used in another part (e.g., body) such that the two fonts used together look visually pleasing. There are three main challenges in font pairing. First, this is a fine-grained problem, in which the subtle distinctions between fonts may be important. Second, rules and conventions of font pairing given by human experts are difficult to formalize. Third, font pairing is an asymmetric problem in that the roles played by header and body fonts are not interchangeable. To address these challenges, we propose automatic font pairing through learning visual relationships from large-scale human-generated font pairs. We introduce a new database for font pairing constructed from millions of PDF documents available on the Internet. We propose two font pairing algorithms: dual-space $k$-NN and asymmetric similarity metric learning (ASML). These two methods automatically learn fine-grained relationships from large-scale data. We also investigate several baseline methods based on the rules from professional designers. Experiments and user studies demonstrate the effectiveness of our proposed dataset and methods.
\end{abstract}

}

\maketitle

\IEEEdisplaynontitleabstractindextext

%
\IEEEpeerreviewmaketitle

\IEEEraisesectionheading{\section{Introduction}\label{sec:introduction}}

\begin{figure}[t]
            \centering
                \includegraphics[width=0.95\linewidth]{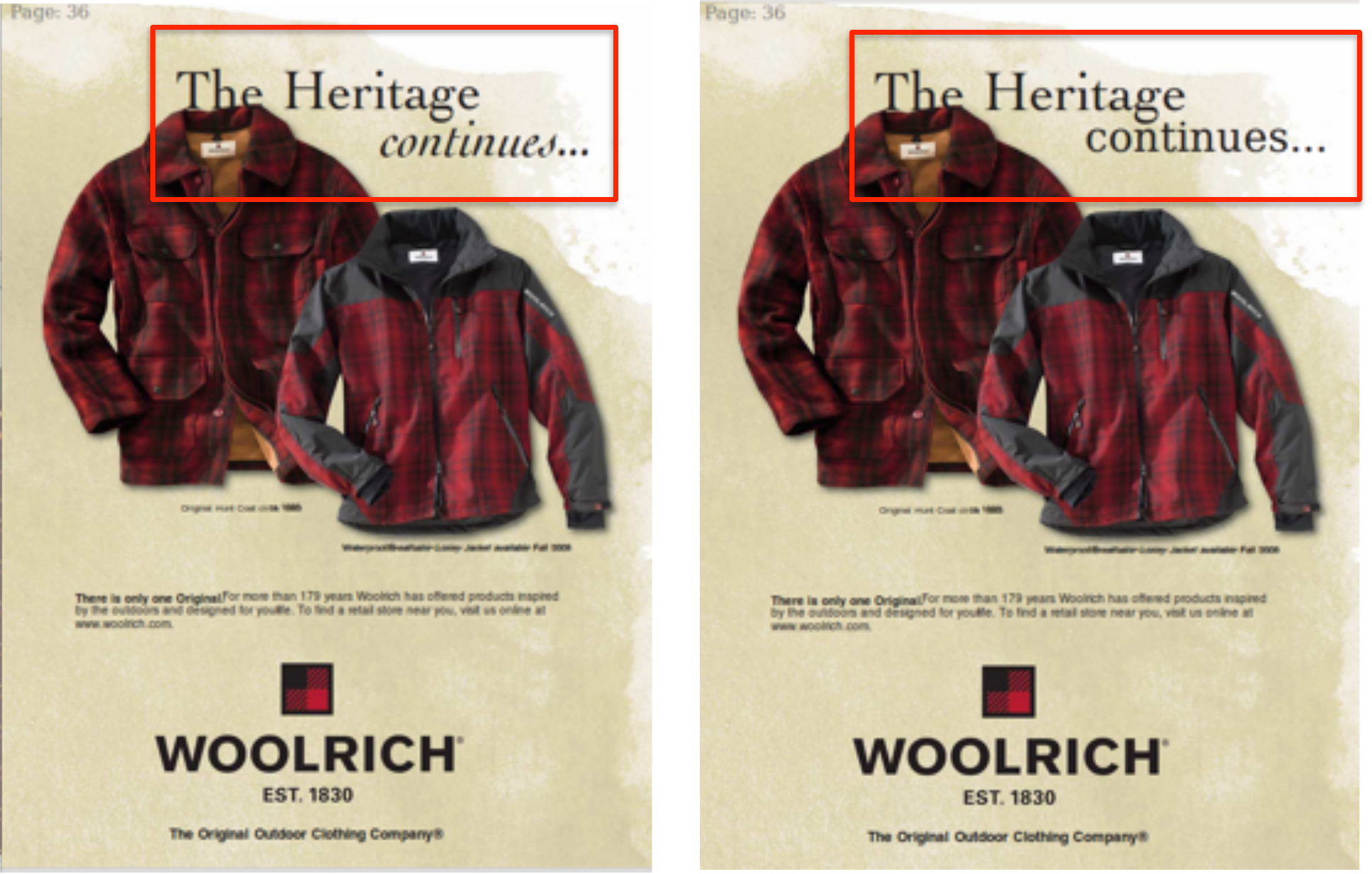}
            \caption{Examples for PDF pages with different font pairing. The font of headers are the same, while the fonts of sub-headers are different. We show the same PDF page rendered with two choices of font for the sub-header (``Continues"), and the same font for the header (``The Heritage'').
            }
            \label{fig:userstudy}
\end{figure}

\IEEEPARstart{I}{}n multimedia filed, applying artificial intelligence to facilitate art and design has drawn a lot of attention recently, such as automatic generation of visual-textural presentation layout \cite{yang2016automatic}, font recognition, election and prediction \cite{o2014exploratory,wang2015deepfont,wang2015deepfontlong,ZhaoPG2018} and visual document analysis \cite{kembhavi2016diagram,siegel2016figureseer}.
Pairing fonts is an important task in graphic design for documents, posters, logos, advertisements and many other types of design. A designer typically picks a title font, sub-header fonts, body text fonts, and does so in a way that is harmonious and appropriate for the style. Fonts should complement each other without ``clashing'' or appearing disconnected.
For example, Figure \ref{fig:userstudy} shows the same advertisement with two choices of font for the sub-header (``Continues"), and the same font for the header (``The Heritage"). Each pair of choices conveys a different design quality: in one case, the fonts complement each other and appear more interesting, whereas when they are nearly the same the layout is less appealing. Each pair of header and sub-header fonts represents a \emph{font pair}. Despite the importance of font pairing, current design tools do not provide much assistance in the challenging task for pairing fonts, aside from providing a few template designs.

Design books and websites provide many rules-of-thumb for font pairing, such as ``Use Fonts from the Same Family'', ``Mix Serifs and Sans Serifs'' and ``Create Contrast''\footnote{{https://designschool.canva.com/blog/combining-fonts-10-must-know-tips-from-a-designer/}}$^{,}$\footnote{{https://webdesign.tutsplus.com/articles/a-beginners-guide-to-pairing-fonts--webdesign-5706}}, but such rules can be hard to apply in practice or to formalize. 
Font pairing is especially challenging for novices creating designs, who may lack the intuitions for selecting fonts.

This paper introduces the problem of automatic font pairing. 
Given a font selection in a document, our goal is to recommend matched fonts that produce pleasing visual effect when they are used together in different parts of a document. For example, given a header font, recommend a body font, or vice versa.  There are three main challenges in font pairing. First, this is a fine-grained problem, which means that subtle distinctions between fonts may be important, as opposed to object-level co-occurrence problems (e.g., sky and airplane) \cite{galleguillos2008object,ladicky2010graph,feng2016semantic} or category-level pairing (e.g., tops and skirts \cite{mcauley2015image,veit2015learning,mcauley2015inferring,jagadeesh2014large,liu2012hi,yu2012dressup}). Second, designers have listed many rules for font pairing, but they are difficult to formalize.  Font pairing is not simply a problem of similarity: designers typically pair contrasting fonts as well as similar fonts. Third, font pairing is an asymmetric problem. Pairing font A as header and font B as body is different as pairing font B as header and font A as body.

{It should be noted that font pairing is a complex task: a good font combination is decided by many factors beyond font itself, e.g., text sentiment, layout and even personal taste. As a first attempt to attack this problem, the goal in this paper is to recommend font pairs that satisfy majority users' aesthetics only based on visual font information. We believe this is less challenging than also considering other elements in design context. It is also a well-posed problem since most tutorials and books \cite{bonneville2010bigbook} on this topic recommend font pairs in the same setting.}

To address these challenges, we propose to learn font pairing from large-scale human-designed font pairs. However, obtaining appropriate data for this problem is challenging, as there is no existing dataset and the font pairs from Internet web pages are noisy or biased to a small set of popular fonts.
We collected a new database called ``FontPairing'' from millions of PDF pages on the Internet. These PDFs embody a very diverse set of designs with font meta data embedded. We devise some heuristics to automatically identify header/sub-header and header/body pairs from the PDF pages with a high accuracy verified on a subset.


Given this data, we investigate two algorithms to learn font pairing: dual-space $k$-NN and asymmetric similarity metric learning method. The intuition behind dual-space $k$-NN is that the users may choose the same body for similar header fonts, and vice-versa. For example, if the font Univers is similar to Helvetica, then fonts that pair well with Univers should pair well with Helvetica.
Given an input query font (e.g., header), we first find the $k$ most similar fonts from the training header fonts. We then rank their corresponding body fonts by their appearance frequency in the training data. Appearance similarity is measured by a deep neural network trained for this task.


The goal of asymmetric similarity metric learning is to learn a distance function by which fonts that pair well have small distances to each other, and, conversely, mismatched fonts are far apart. The metric is discriminatively trained from our training font pairs. Especially, we jointly learn the model that bridges the asymmetric similarity and distance metric. At test time, an online prediction entails finding the nearest font pairs in the dataset.

To the best of our knowledge, our work is the first to address the automatic font pairing problem. Since there is no prior work, we compare with several baseline methods (e.g., same font family, similarity, contrast) according to the rules provided by professional designers. Experimental results show effectiveness of our visual learning based dual-space $k$NN and asymmetric similarity metric learning methods.

\section{Related Work}

\begin{figure*}[t]
            \centering
                \includegraphics[width=0.95\linewidth]{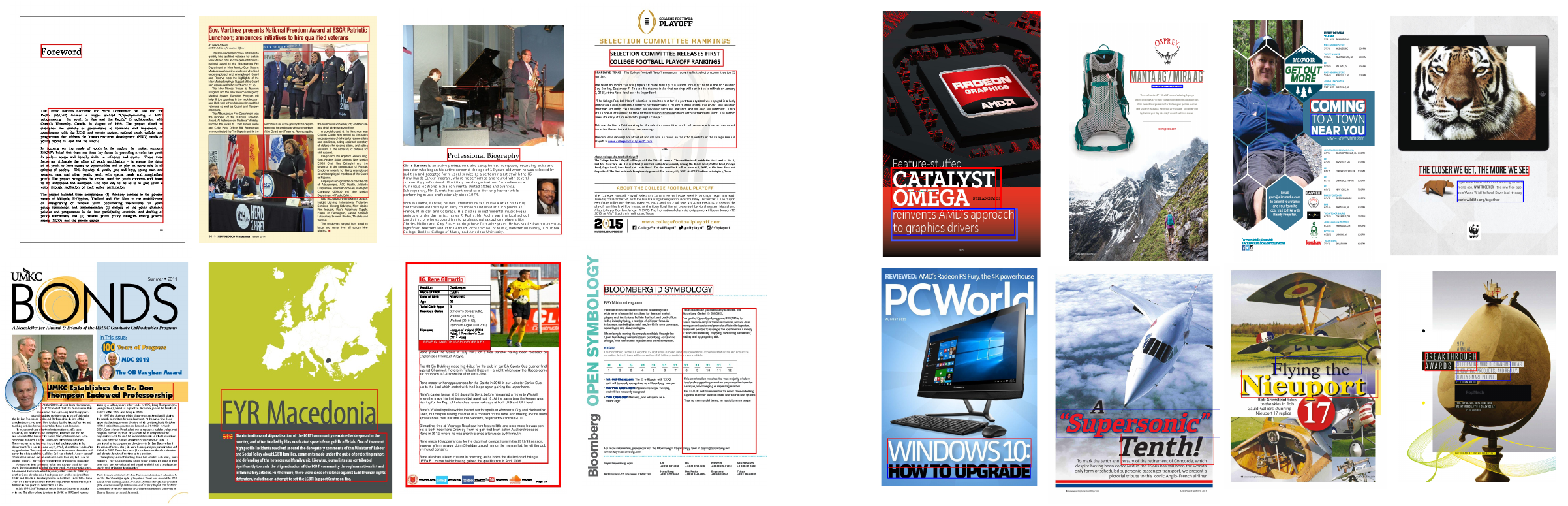}
            \caption{Examples for font pair detection in freely-available PDFs. The  left four columns show documents with detected header/body pairs, marked with red bounding boxes.
            The right four columns show examples of documents with detected header/sub-header pairs, with headers marked in red and sub-headers in blue.
            \label{fig:PDFs}
            }
\end{figure*}




To the best of our knowledge, our work is the first to address font pairing problem, and there is very little related work.

In multimedia and computer vision fields, several methods have been proposed for font recognition 
\cite{chen2014large,wang2015real,wang2015deepfont,wang2015deepfontlong} and font prediction \cite{ZhaoPG2018} based on large datasets of fonts and their images. Wang et al.~\cite{wang2015deepfont,wang2015deepfontlong} train deep neural networks for font recognition. Zhao et al. \cite{ZhaoPG2018} proposed multi-task deep neural networks to jointly predict font face, color and size for each text element on a web design, by considering multi-scale visual features and semantic tags of the web design. Our work is also related to systems for learning to parse web pages, such as WebZeitGeist \cite{webzeitgeist}. The most relevant to our work is by O'Donovan et al.~\cite{o2014exploratory}, who present interfaces for finding fonts based on learned models of font style. However, their work focuses only on single fonts in isolation, whereas we consider how two fonts pair with each other. Font pairing is also related to visual document analysis (e.g., \cite{kembhavi2016diagram,siegel2016figureseer}) and automatic generation of visual-textural presentation layout \cite{yang2016automatic}.

In terms of methodology, our work is highly related to other visual pairing tasks, particularly pairing clothing \cite{mcauley2015image,veit2015learning,mcauley2015inferring,jagadeesh2014large,liu2012hi,yu2012dressup}, furniture \cite{liu2015style}, and food \cite{Ahn2011}.
Here we address font pairing, which entails particular difficulties including lack of an appropriate data source, fine-grained difference between font types and symmetrically pairing entities of the same category instead of different categories.

\section{A Database for Font Pairing}
In this section, we introduce the new database we generated for visual font pairing task called ``FontPairing''. We collect millions of freely-available PDFs on Internet, analyze and extract header/body font pairs and  header/sub-header pairs from the PDF pages.


\subsection{Font Pairings From Web}

Perhaps the most obvious approach to gather font pairing data is to obtain them from webpages such as Google Fonts\footnote{https://fonts.google.com}, Typ.io\footnote{http://andreasweis.com/webfontblender/}, Typewolf\footnote{https://www.typewolf.com}. For example, each font on Google Fonts is provided with a list of suggested pairings. However, we found these datasets inadequate for two reasons. First, they each provide a small set of pairings. The second and more significant problem is that these pairing lists are extremely unbalanced: these websites generally recommend only popular body fonts. Out of the above pairs, 43\% of the font pairs involve one of the five most-popular fonts.

\subsection{PDF Dataset}

To address these problems, we propose to detect font pairs from millions of PDF documents available on the Internet. We have collected more than 300,000 PDF files from various websites such as Commons.wikimedia.org and Digital-library.usma.edu. Each PDF usually contains dozens to hundreds of pages; from all these PDFs, we obtain more than 15 million pages in total.  As shown in Figure \ref{fig:PDFs}, these PDF pages exhibit various layouts, topics, and font styles. 
We believe this dataset could be potentially useful for training other models for document design as well.


A key challenge is then to extract visual information from this large dataset. Although PDF is a structured document format, it is complex and does not include the annotations we need (e.g.,``header font"). Parsing such structured representations is a major challenge in itself \cite{webzeitgeist}. Rather than attempting to fully parse the document, we focus only on identifying the font pairs containing the header, sub-header, and/or body fonts.

Of the PDFs we collected, 43\% are scanned documents. We omit these from the dataset, to simplify parsing and avoid additional noise caused by the parsing processing. 
For the remaining data, we apply open PDF tools to extract text, image, and layout information from each page of each document. 
We define a \textit{text box} as a several words with the same font style and size in a line.
Each text box is annotated with the font style, font size, and the bounding box. We discard pages that contain fewer than two text boxes. We also focus only on pages with the Roman alphabet, which we identify using  Python language detection tools. Of the dataset, 75\% of documents are in English. 


To detect a header and sub-header pair on a page, we first find the largest text box, and call this the header text. We then identify the largest text box that lies within a fixed threshold of the header text box. We then call this a header/sub-header pair, and extract the fonts from the two text boxes. Only one header/sub-header pair is found for each document. 
We also detect body text boxes by finding text boxes with number of characters above a threshold. The nearest body text box to a header is used to form a header/body pair.
Figure \ref{fig:PDFs} shows example detection results on both header/body and header/sub-header pairs.


To evaluate the accuracy of automatic pair detection on PDF, we manually label header/body and header/sub-header pairs on a small subset of PDFs (i.e., 20 PDFs with varies topics and layouts totalling 3,000 pages). Here, our purpose is not to evaluate whether these are good pairing or not. We manually compare the automatic detection results with human labeling for verification. By adjusting our detection thresholds (e.g., the distance between the text boxes of header and sub-header), we achieve about 95\% precision (true positives) in our automatic detection. For header/subheader pairs, our detector achieves 85\% precision (true positives).
There are more variations in the layout of header/subheader, which makes this task much harder than detecting header/body pairs.



\begin{figure}[t]
            \centering
                \includegraphics[width=0.93\linewidth]{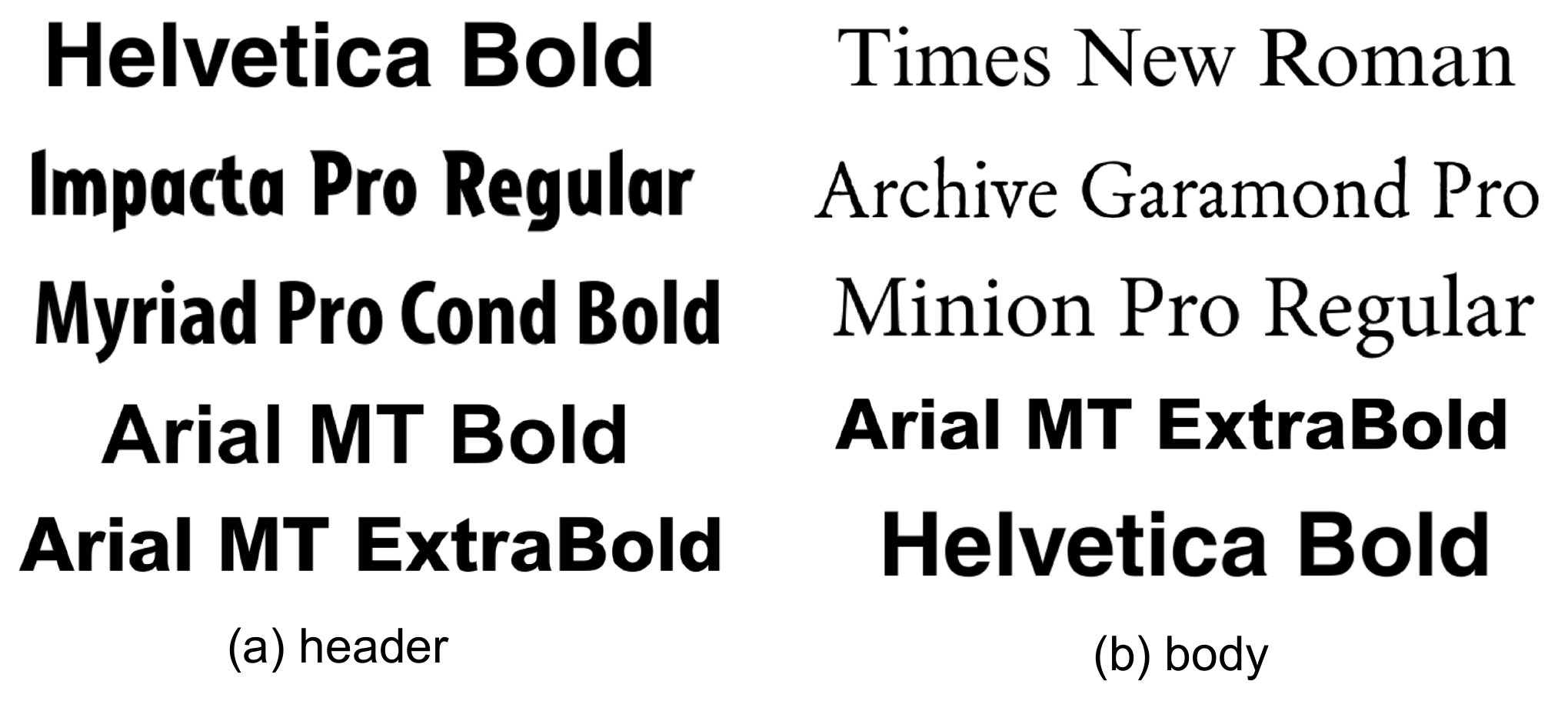}
            \caption{Top 5 header and body fonts used in header/body pairs.  }
            \label{fig:fontstop}
\end{figure}
\begin{figure}[t]
            \centering
                \includegraphics[width=0.98\linewidth]{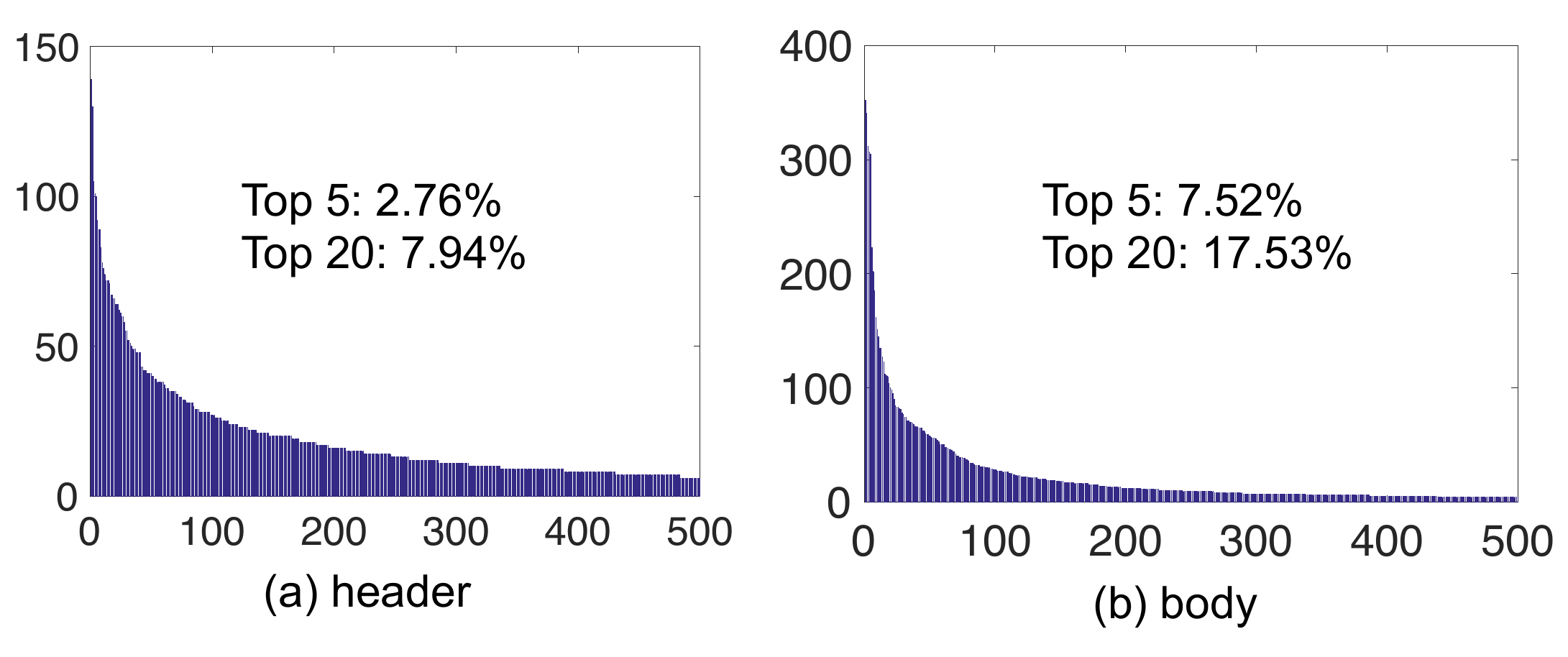}
            \caption{Data distribution of head/body pairs in FontPairing dataset. The histogram of the number of times a header font appears in unique font pairs is shown in (a) and the histogram of a body font is shown in (b).
            Fonts are written in the PostScript font name format, which typically includes both font family (e.g., ``Helvetica") and style (e.g., ``Bold").
            \label{fig:wikidis} }
\end{figure}

{The number of total unique header fonts, sub-header fonts, body fonts and pairs are shown in Table \ref{Table:Tab07}.}
Figure \ref{fig:fontstop} shows the top 5 header and body fonts used in header/body pairs and Figure \ref{fig:wikidis} shows the histogram of frequency a header or body font appears in unique font pairs. Only 2.7\% of header/body pairs involves one of the five most popular header fonts, and 7.5\% of pairs involve one of the five most popular body fonts. 
This indicates a far more diverse set of pairings than web recommended pairings, of which, as reported above, 43\% involve one of the five most-popular body fonts.

\begin{figure}[t]
            \centering
                \includegraphics[width=0.93\linewidth]{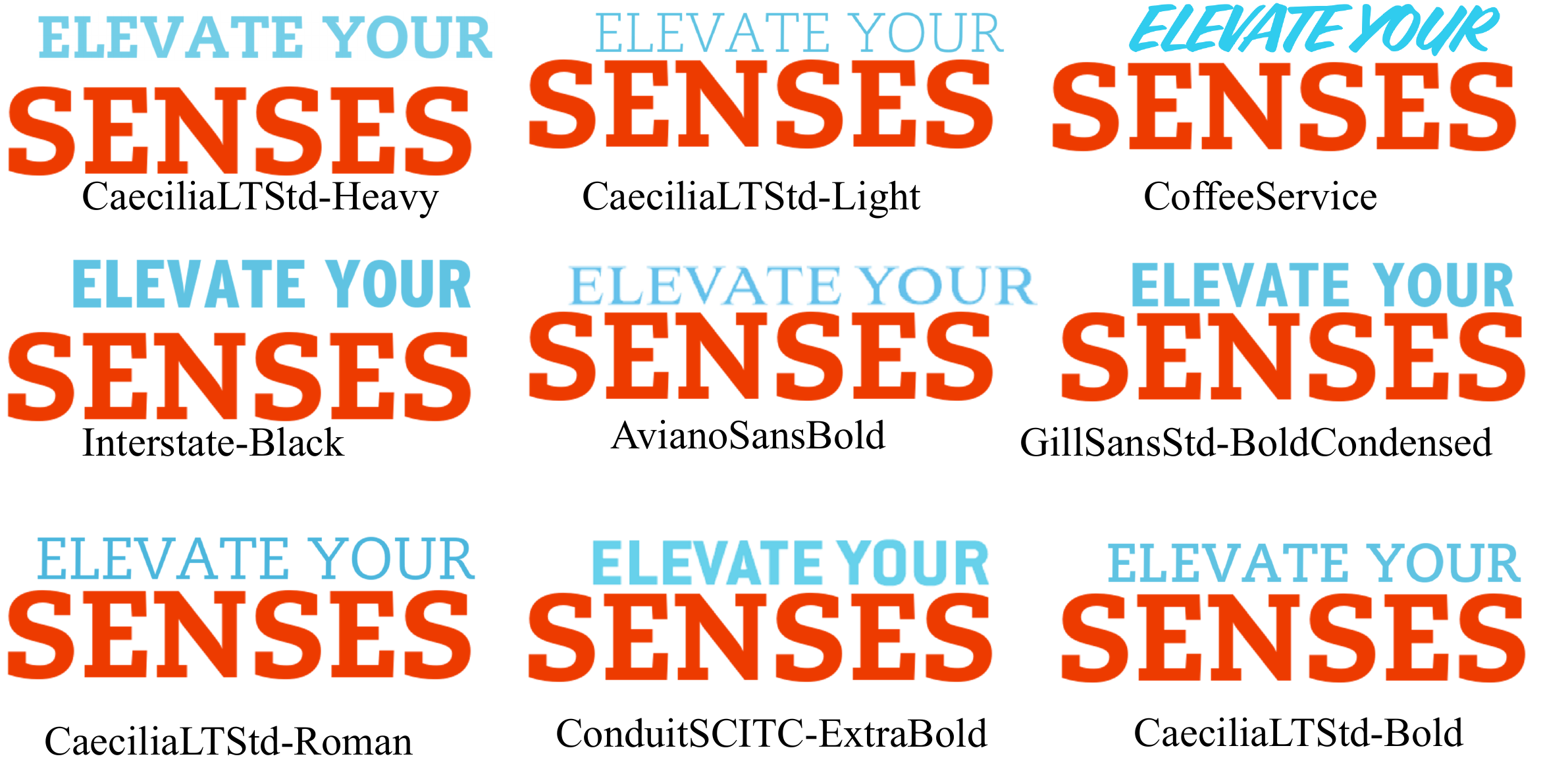}
            \caption{Examples of head/sub-header pairing in FontPairing dataset, for the header font ``CaeciliaLTStd-Heavy".  }
            \label{fig:fontexample}
\end{figure}

Figure \ref{fig:fontexample} shows sample font pairings in our dataset when given a query header font ``CaeciliaLTStd-Heavy".  For this font, our datset includes 10 font pairs, and 20 for the entire ``Caecilia'' family (including Bold, Heavy, Italic, etc).
This is much more diverse than those in web font sources.
For example, for this same font, there are only 4 header/sub-header pairings on Fontinuse.com, 1 pairing on typewolf.com, 2 on typ.io.com, and 0 on Google Fonts.

\subsection{Quality Verification}


Following Veit et al.~\cite{veit2015learning}, we conduct an online user study to compare whether designers and ordinary users prefer the real font pairs we detected from PDFs or the random alternatives. 
{Our study includes 60 participants: 15 experts in graphic design (either students in art design major or staff in design company) from Upwork \footnote{www.upwork.com}, and 45 non-designers with other backgrounds from Amazon Mechanical Turk \footnote{www.mturk.com/mturk/welcome} and volunteers.}

The study comprised a set of paired comparisons. In each comparison, a user is shown two images of the same layout, but with one font changed (Figure \ref{fig:userstudy}). In particular, either the header or sub-header font is replaced by selecting a random alternative. The viewer is then asked which design they prefer. We perform two variants of the study: in the first one, the entire page layouts are shown to the viewer; in the second, the user is only shown part of the page containing the relevant text boxes, so that they will focus more on the font choices rather than the context. In whole-page setting, we show 20 comparisons to the users, and in sub-page setting, we show 50 comparisons to the users. These samples are randomly sampled from all the pairs.

Under both whole-page and sub-page settings, experts prefer the original layout 75\% of the time. Non-experts prefer the original 65\% of the time when viewing the full page, and 60\% of the time when viewing the sub-page. Note that the original layout is not necessarily superior to the font choice in the random selection, for various reasons; however, we would expect that it would be more likely to be better. Hence, these results indicate that the pairing combinations included in the dataset are aligned with the preference of expert and common users.
These results also suggest that non-experts are much less sensitive to good font choices than experts, and that there is potential value to recommend good pairings to them.

 We would like to clarify that, although the font pairs extracted from PDF dataset are of varying quality, by training on a large amount of data, we aim to smooth out the noise therein and discover the general pairing rules that match majority users' preference.

\begin{table}[!t] \small
\centering \caption{{Number of unique fonts and pairs of header/body pairing (upper) and header/sub-header pairing (bottom) are shown in the ``full'' column. Number after removing pairs with top 50 famous body/sub-header fonts are shown in the ``non-popular'' column.}}
\label{Table:Tab07} {
 \renewcommand{\arraystretch}{1.1}
    \begin{tabular}{lcc}
     \Xhline{1pt}

font set & full & non-popular \\

   \Xhline{1pt}  
unique headers &2,086&616 \\ 

unique bodies &1,443&1,343\\


header/body pairs &13,251&5,337\\ 
  \Xhline{1pt}

unique headers &2,159&1,054 \\ 

unique sub-headers &2,168&1,573\\


header/sub-header pairs&8,733&5,174\\ 

\Xhline{1pt} 
\end{tabular}
}
\end{table}

\section{Methods}

Given a dataset of font pairs, our goal is to learn a model for predicting good font pairs. For example, given a header font, we would like to recommend good body fonts to go with it. We learn separate models for header/body and header/sub-header pairings. Without loss of generality, we discuss the header/body pairing in the rest of the section.

Suppose we have $m$ training header fonts with feature vectors $\{\bx_1, ..., \bx_m\}$, and $n$ training body fonts with features $\{\by_1, ..., \by_n\}$. For each header font $\bx_i$, there is a list of body fonts that pair with it, i.e., $\mathcal{P}_i = \{\by_{i1}, \by_{i2}, ...\}$. Fonts may repeat in this list, so that the popularity of pairings can be captured in the data.

\begin{figure}[t]
            \centering
                \includegraphics[width=0.8\linewidth]{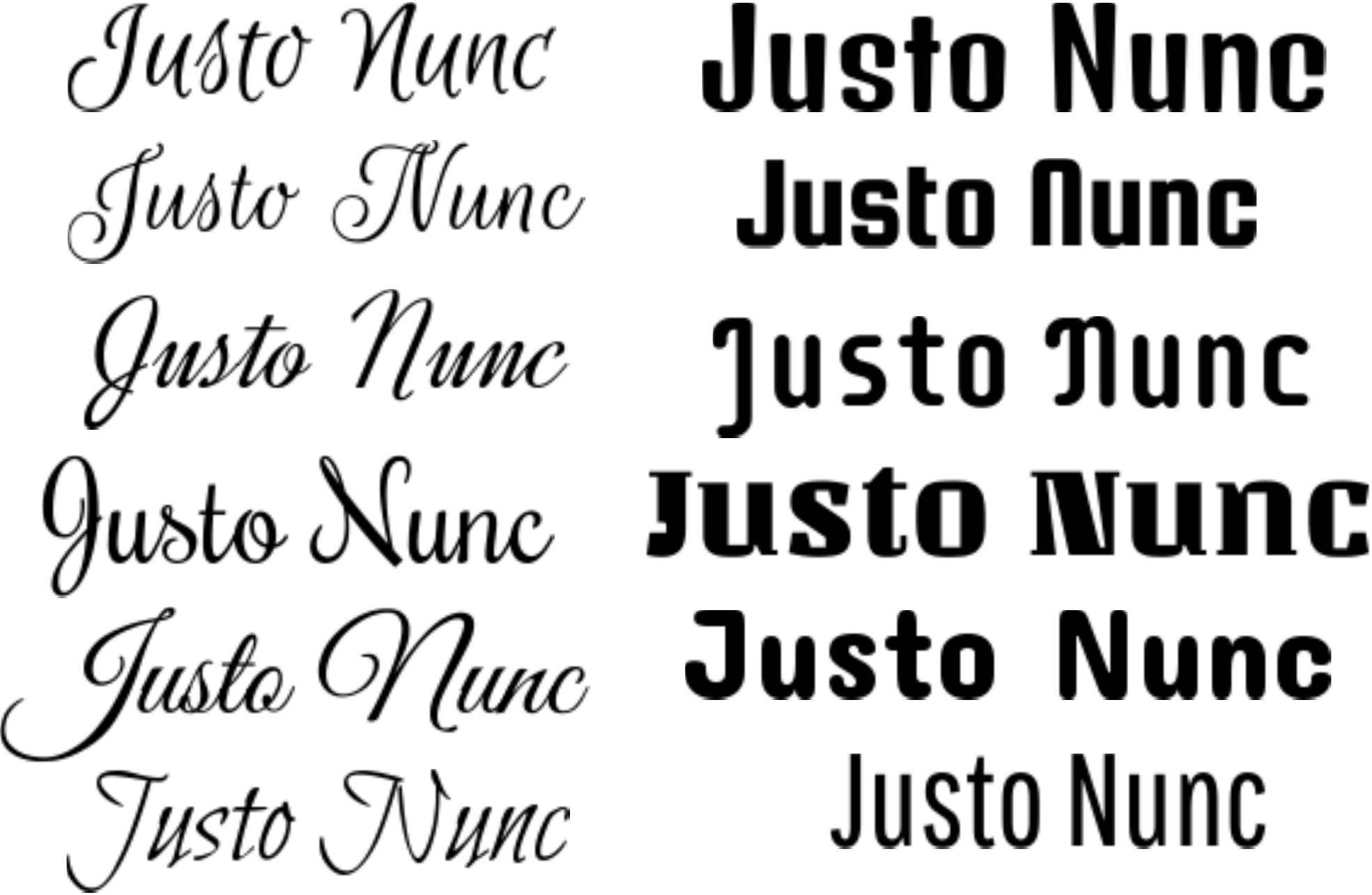}
            \caption{Two examples for similar font retrieval based on DeepFont features. In each column, the first row is the input query font. The following rows present top 5 similar fonts measured with the distance of DeepFont feature. The robustness of DeepFont features facilitate the performance of dual-space $k$NN method.
            \label{fig:similarity} }
\end{figure}

 We use pretrained feature of each font from DeepFont method \cite{wang2015deepfont,wang2015deepfontlong} as the input font feature representation. DeepFont model is trained for font recognition on the large-scale Visual Font Recognition (VFR) dataset. DeepFont introduces a Convolutional Neural Network (CNN) decomposition approach, using a domain adaptation technique based on a Stacked Convolutional Auto-Encoder (SCAE) that exploits a large corpus of unlabeled real-world text images combined with synthetic data. Using this model, we obtain the feature vector for a font $i$ denoted as $\bx_i \in R^D$, where $D=768$. As shown in Figure \ref{fig:similarity}, distances in the DeepFont feature space correspond to perceptual similarity between fonts. It demonstrates the effectiveness DeepFont feature for searching for perceptually-similar fonts. We do not choose to apply the end-to-end deep neural network (DNN) to learn font pairing. The main reason is that the number of unique header/body pairs and header/sub-header pairs in our database is 13,251 and 8,733 respectively as shown in Table \ref{Table:Tab07}, which is not enough to train an end-to-end DNN.


In the following, we discuss two methods designed for font pairing: dual-space $k$-NN (DS-$k$NN) and asymmetric similarity metric learning (ASML).

\subsection{Dual Space $k$-NN Search based Method}


The intuition behind dual space $k$-NN search based method (DS-$k$NN) is that, if fonts $F_1$ and $F_2$ are similar, then fonts that pair with $F_1$ should be good pairings with $F_2$.

Suppose we are querying which body font will go with a header font $\bx_q$.
We first find the top $K_1$ nearest header fonts $[\bx_1',...,\bx_i',...,\bx_{K_1}']$, based on cosine similarity ${\rm cos}(\bx_q,\bx_i)$ in feature space between $\bx_q$ and all the training headers $\{\bx_1,...,\bx_i,..., \bx_m\}$. Each header $\bx_i'$ has a list of body fonts that pair with it, i.e., $\mathcal{P}_i' = \{\by_{i1}', \by_{i2}', ...\}$. The fonts in $ \tilde{\mathcal{P}}  = \{\mathcal{P}_1',...,\mathcal{P}_i',...,\mathcal{P}_{K_1}'\} $ are regarded as candidate body fonts for pairing $\bx_q$. We assume that there are $N_1$ fonts in candidate body font set $\tilde{\mathcal{P}}$. Note that fonts may repeat in this list. A high frequency of repeat in this list demonstrates that in the training set, more similar headers are paired with this body font.   


The candidate body fonts may only cover a part of the good pairings among all the body fonts. Fonts similar to the candidate body fonts may also result in pleasing pairings. Therefore, we rank all the $n$ body fonts $\{\by_1,...,\by_j, ..., \by_n\}$ based on the similarity score $\tilde{S}(\by_j)$ compared with candidate body fonts, and recommend top $N$ fonts with highest scores.

Here we introduce the way to calculate $\tilde{S}(\by_j)$ for font $\by_j$. We first calculate the cosine similarity ${\rm cos}(\by_j,\by_l)$ between $\by_j$ and each candidate body font $\by_l \in \tilde{\mathcal{P}}$. Second we select top $K_2$ candidate body fonts with the largest ${\rm cos}(\by_j,\by_l)$. Then we calculate the average of cosine similarity ${\rm cos}(\by_j,\by_l) (l \in\{1,...,K_2\})$ by multiplying ${\rm cos}(\bx_q,\bx_l)$, which calculates the similarity of similar header $\bx_l$ ($\by_l \in \mathcal{P}_l'$ ) and query header $\bx_q$: 
\begin{equation}\label{Eq:Eqdsknntf}  
\tilde{S}({\by_j}) = \frac{1}{K_2}  \sum_{l=1}^{K_2} {\rm cos}(\by_j,\by_l)  \cdot {\rm cos}(\bx_q,\bx_l).
\end{equation} 

 Note that fonts may also repeat in this list of top $K_2$ candidate body fonts. It is similar as the idea of adding a tf weight ${\rm {tf} (\by_l)}$ in tf-idf (short for term frequency-inverse document frequency \cite{aizawa2003information}) for each unique font in $K_2$ candidate body fonts. In this way, the fonts with a high frequency in the list are assigned with higher weights. The idf weight could be further integrated in Eq. (\ref{Eq:Eqdsknntf}) by multiplying ${{\rm idf} {(\by_l)}}$ for each $\by_l$ as:
\begin{equation}\label{Eq:Eqdsknntfidf}  
\hat S({\by_j}) = \frac{1}{K_2}  \sum_{l=1}^{K_2} {\rm cos}(\by_j,\by_l)  \cdot {\rm cos}(\bx_q,\bx_l) \cdot{{\rm idf} {(\by_l)}}, 
\end{equation}where ${{\rm idf} {(\by_l)}}=\frac{m}{{t_l}}$ and ${t_l}$ is the number of header fonts with which body font $\by_l$ is paired in the training set. The main purpose for adding the idf weight is to reduce the impact of popular body fonts.

{However, DS-$k$NN does not perform well if accurate similar header fonts are hard to find in the dataset. Also, popular body fonts have more chances of pairing similar header fonts and appearing in the set of candidate body fonts in this method. Meanwhile, there are some font pairing rules that may be missed by DS-$k$NN. For example, using the same font family for the header and body (e.g., Helvetica Bold for header and Helvetica for body). These rules are difficult to capture and could not be easily solved by calculating font similarity in the original feature space. These concerns motivate us to learn the metric in the next section to capture the common pairing strategies.}



\subsection{Asymmetric Similarity Metric Learning}

The goal of Asymmetric Similarity Metric Learning method is to learn a better distance scoring function between fonts, so that fonts that pair well have low distance, and mismatched fonts have large distance.  We train this scoring function offline. Then predictions are generated for a given query by finding the fonts with lowest distance based on new scoring function.


We treat the training dataset as comprising font pairs $(\bx_i, \by_j)$, and an indicator function $S(i,j)=1$ when fonts are paired in the training dataset. Since our FontPairing dataset only containing positive pairs, we randomly sample negative pairs among all the other possible pairs excluded these positive pairs. The number of negative pairs and positive pairs are the same. The indicator function $D(i,j)=1$ when fonts are negative pairs. While there may exist good pairings among the negative set, but these should be in the minority, especially since the user study found that positive pairs were more attractive than randomly picked pairs to designers. Generally speaking, the original font pairs in PDFs are usually specific designed and should achieve higher accordance than randomly picked ones. 



The main idea for conventional metric learning \cite{weinberger2009distance} is to learn the a better scoring function $||\bx - \by||_\bM^2 = (\bx-\by)^\intercal\bM(\bx-\by)$, to enlarge the two font points of non-matching pairs and narrow the font points for matched pairs.   
The learning objective function is:
\begin{equation}\label{Eq:Eq01}  
\begin{array}{l}
\mathop {\min }\limits_\bM \sum\limits_{i,j}^{m,n} {||{\bx_i} - {\by_j}||_\bM^2}  \cdot {S_{ij}}\\
s.t.\;\bM \succ  = 0,\;    \mathop \sum\limits_{i,j}^{m,n} {||{\bx_i} - {\by_j}||_\bM^2}  \cdot {D_{ij}} \geq 1.
\end{array}
\end{equation}


Although metric learning is very important for many supervised learning application (e.g., classification), it has a few limitations in our font pairing problem. First, instead of applying nearest neighbor classifiers with ML in classification problem, after ML, we still need to make a decision, such as with a constant threshold $d$:
\begin{equation}\label{Eq:Eq04} 
f_{ML}(\bx_i,\by_j) = d - (\bx-\by)^\intercal\bM(\bx-\by).
\end{equation}

However, a simple constant threshold may be sub-optimal, even if the associated metric is correct. Another challenge is that, font pairing is asymmetric. It means paring A as header and B as body is
different as pairing B as header and A as body. To address these challenges, we consider a jointly model that bridges a learn a distance metric and a asymmetric similarity decision rule and propose a asymmetric similarity metric learning as:
\begin{equation}\label{Eq:Eq05} 
f_{(\bM, \bG)}(\bx_i,\by_j) = \bx^\intercal \bG \by - (\bx-\by)^\intercal\bM(\bx-\by),
\end{equation}where $\bG$ is asymmetric. $\bx^\intercal \bG \by $ measures the similarity of font pairs.

Let $P = S\cup D$ denotes the index set of all pairwise constraints. Let $y_{i,j} = 1$ if $S (i,j) = 1$ and $y_{i,j} = -1$ if $D (i,j) = 1$. We drive the formulation of the empirical discrimination using hinge loss:
\begin{equation}\label{Eq:Eq06} 
\begin{array}{l}
\mathop {\min }\limits_{\bM, \bG}\sum\limits_{(i,j)\in P} {(1-y_{i,j} f_{(\bM, \bG)}(\bx_i,\by_j) )_+}\\
+ \gamma /2 (\|\bM-I\|^2_F + \|\bG-I\|^2_F ),
\end{array}
\end{equation}where the regularization term $\|\bM-I\|^2_F + \|\bG-I\|^2_F$ prevents the image vector being distorted too much. $\|\cdot\|_F$ is the frobenius norm. $\gamma$ is the trade-off parameter. This objective function could be solved with dual formulation as \cite{cao2013similarity}.


After off-line learning the new scoring function as Eq. (\ref{Eq:Eq05}), in online pairing, we recommend font pairs according to the distance between header and body based on the new scoring function. 


\section{Experiments}


\subsection{Compared Methods}

We implement the following baselines for comparison, including several based on design rules-of-thumb.



\noindent \textbf{Popularity}: The method aims at recommending most popular body fonts. First, we rank all body fonts according to the frequency they appear in font pairs in the collected dataset. The top-ranked body fonts are defined as popular fonts. These same fonts are always recommended, regardless of the query header.




\noindent \textbf{Simple $k$NN (S-$k$NN)}: This method aims as recommending body fonts with highest visual similarity to the query header font as pairs. The distance of similarity between to fonts is measured by DeepFont features.


\noindent {\textbf{Contrast similarity (ConSim)}: The main intuition in this work is that the ideal pairing has similarities and contrasts in equal importance. They manually designed a contrast similarity distance metric. More details could be found at \cite{consim}.}




\noindent \textbf{Similarity metric learning (SML)}: We also implement a similarity metric learning method (SML) to evaluation the effectiveness of making the metric $\bG$ asymmetric in ASML. We replace asymmetric $\bG$ in Eq.(\ref{Eq:Eq05}) and (\ref{Eq:Eq06}) with symmetric metric. This idea is similar as \cite{cao2013similarity}, but \cite{cao2013similarity} address on face verification problem.


\noindent \textbf{Dual-space $k$NN (DS-$k$NN)(Ours)}:  Our proposed dual space $k$NN method. 

\noindent \textbf{Asymmetric similarity metric learning (ASML) (Ours)}: Our proposed asymmetric similarity metric learning method.

\begin{figure*}[t]
            \centering
                \includegraphics[width=0.98\linewidth]{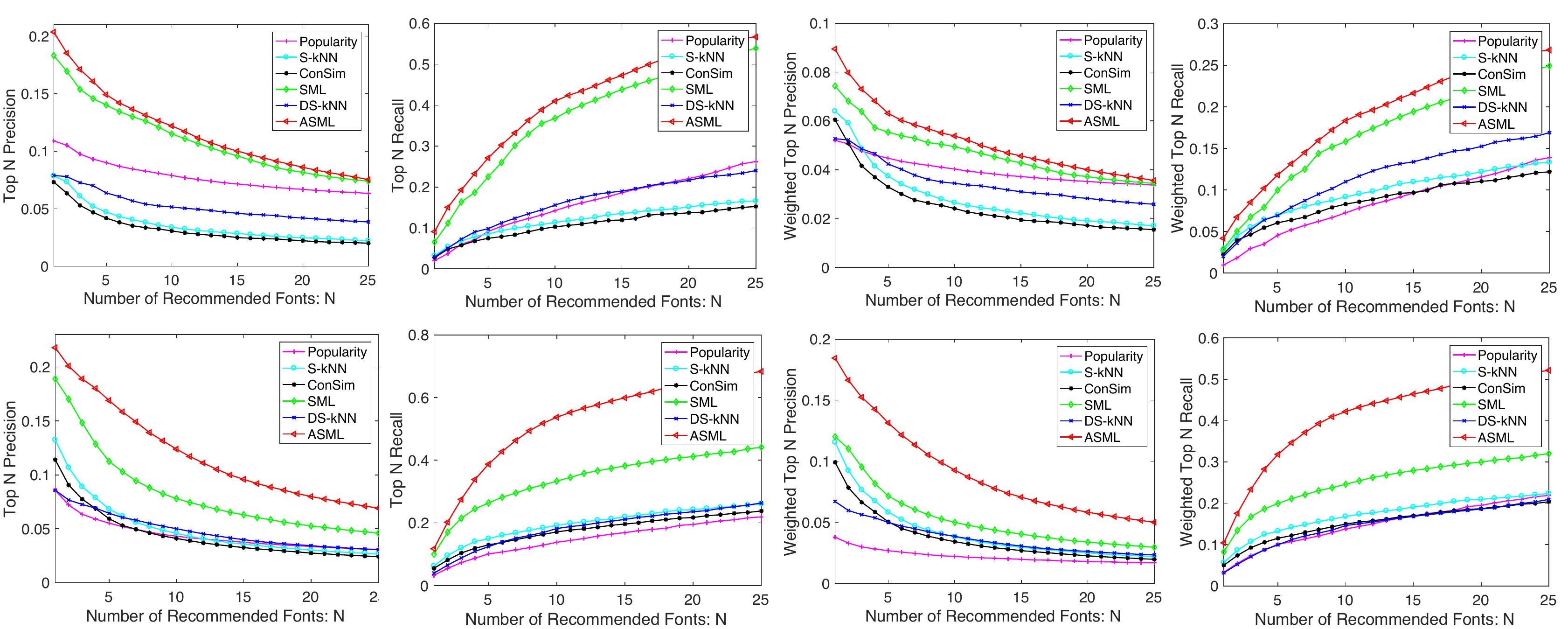}
              \caption{{Performance of top $N$ recommendation on header/body (first row) and header/sub-header (second row) pairing with top $N$ precision and recall and weighted top $N$ precision and recall evaluation metric.}}
            \label{fig:topNall}
\end{figure*}   

\begin{figure*}[t] 
            \centering
                \includegraphics[width=0.96\linewidth]{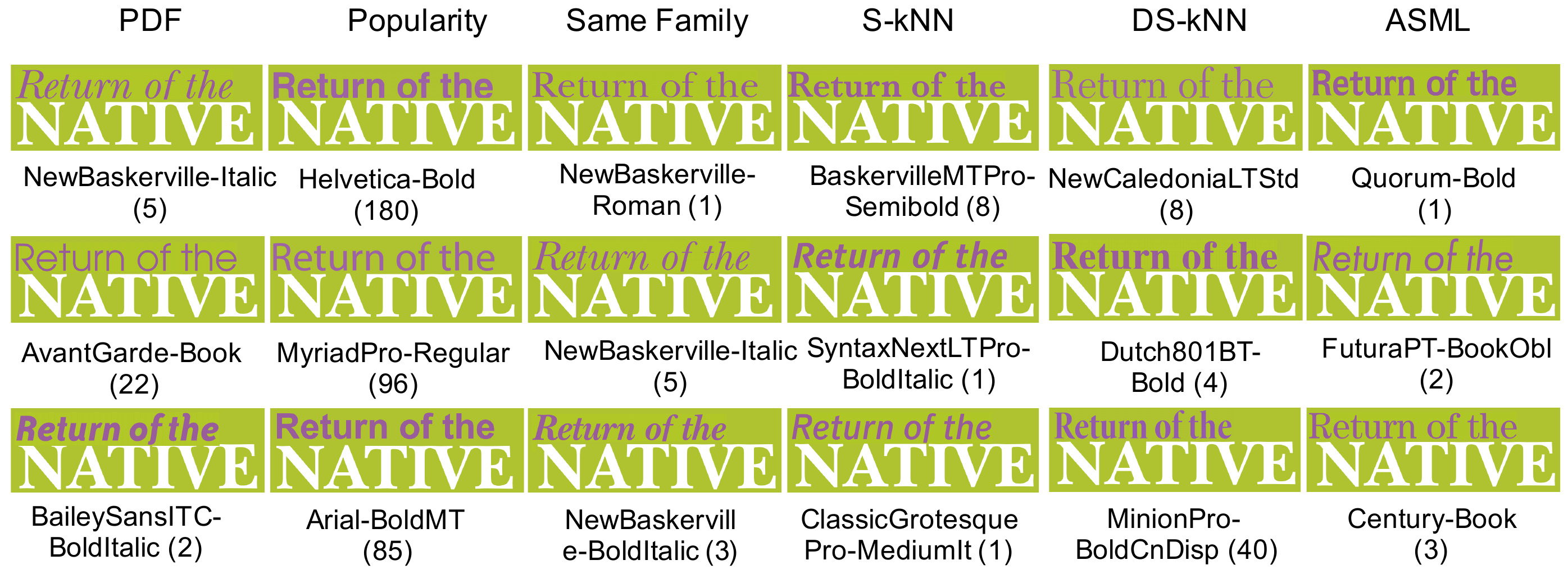}
            \caption{Examples of header/sub-header font pairing results. The input is a query header font ``NewBaskerville-BoldSC''. All the pairs are rendered with same format and header font, but only with different sub-header fonts. The most left column shows three pairings in our collection. In each column, we show the top 3 recommended sub-header fonts by each methods. The PostScript name of each sub-header is shown below the image. The number of times the sub-header font appears in the unique font pairs are shown within the parentheses. We recommend to zoom in to get more details of the fonts.}
            \label{fig:visualization1}
\end{figure*}

\subsection{Experimental Setting}

We perform quantitative evaluation similar to other pairing tasks \cite{veit2015learning,mcauley2015image,jiang2015author}. We conduct two sets of experiments: top-$N$ recommendation and binary classification. Without loss of generality, we discuss the setting that given a header font, we recommend body pairings as an example in the rest of the section. 

The first evaluation is to formalize the pairing problem as a retrieval problem. Given a header font, we rank all the body fonts and recommend top-$N$ body fonts as good pairings. The second evaluation is to formalize the pairing problem as a binary classification problem: given a font pair, we want to classify whether it is a good pairing or not.

We split the header fonts in FontPairing dataset into training header set and test header set by a ratio of 9:1 with no overlap. Only the pairings with training headers are used as positive training pairings. In this way, in the test stage, we are able to evaluate the performance of recommending body fonts to pair an unseen header font. The body fonts in the training and test set may have overlaps.

\subsubsection{Top-$N$ recommendation}
In real world font pairing interfaces, we would like to recommend multiple candidate fonts for pairing, and let the user pick from this list \cite{o2014exploratory}. Thus, we evaluate top-$N$ recommendation performance, namely, precision and recall at $N$, which are widely used in recommender systems \cite{gunawardana2009survey}.

Assuming the user gets a top-$N$ recommended list of fonts, \textbf{recall} is the percentage of relevant fonts selected out of all the ground truth fonts, and \textbf{precision} is the percentage of the $N$ results that are good recommendations. 



{Besides the conventional top-$N$ precision and recall, we also apply \textbf{weighted precision} and \textbf{weighted recall} as the evaluation metric. Popular fonts are easier to be considered and may be less interesting to users. We add an IDF weight \cite{salton1988term} (the popular the lower) to each font as:
\begin{equation}\label{Eq:Eq07} 
\text{weighted\_precision}= \text{weighted\_TP}/N,\\
\end{equation}
\begin{equation}\label{Eq:Eq08} 
\text{weighted\_recall}= \text{weighted\_TP}/\text{weighted\_GT},
\end{equation}where $\text{weighted\_TP}$ is the sum of all the IDF weights of true positive fonts. $\text{weighted\_GT}$ is the sum of all the IDF weights of ground truth fonts.}


\subsubsection{Binary classification}

Following the experimental settings from clothing pairing works \cite{veit2015learning,mcauley2015image}, we formalize evaluation in terms of binary classification. Given a header and body font pair, we want to classify whether or not it is a good pairing.

We regard all the font pairs we extracted from PDFs as positive samples. The test set is formed with positive samples and negative samples of equal proportion. Thus, chance performance is 50\% for all experiments. The negative samples are randomly-sampled pairs, excluding all positive pairs, following \cite{veit2015learning,mcauley2015image}. In ``Quality Verification'' section, we have shown that both designers and average users generally prefer the pairings extracted from PDF documents than random chances. 




\subsection{Top-$N$ Recommendation Results}
{


Figure \ref{fig:topNall} shows the performance of top-$N$ recommendation on header/ body pairing and header/sub-header pairing under 6 methods. We show the performance of each method under two metrics: top-$N$ precision and recall and weighted top-$N$ precision and recall. The number of recommended fonts $N$ is shown in x-axis and the corresponding top-$N$ precision and recall are shown in y-axis. The best results of each method are shown in the figure and the parameters are set based on cross-validation.



In all the cases, ASML achieves the highest performance among 6 methods. It demonstrates the effectiveness of regarding visual font pairing as the asymmetric metric learning problem. Also, ASML outperforms SML, which demonstrates the effectiveness of the asymmetric constraint. ASML could automatically learn various font pairing rules and outperform existing rules such as similarity and contrast similarity.



Figure \ref{fig:visualization1} shows the visualization of top recommended sub-header fonts by comparison methods, given a query header font ``NewBaskerville - BoldSC''. The most left column shows the pairings extracted from PDFs. Popularity method tends to recommend fonts with the highest number of frequency shown in parentheses. Same Family method randomly picks fonts from the same font family.
Generally, the font pairings in PDF include many cases that fonts are from the same font family, since it is easy to implement. While Same Family and Popularity method would hit more of PDF extractions, it is very easy for users to pick font with family by themselves, so that the users especially designers may be not interested in these recommendation. Other concern is that it may fail to find same family font for some less popular fonts. S-$k$NN shows the pairings with the smallest visual distances. In DS-kNN(ours) and ASML(ours) methods, we are able to recommend font pairings that are both interesting and unpopular, and meanwhile achieve the coordination of pairing.




}

\subsection{Binary Classification Results}


Table~\ref{Table:Tab01} shows binary classification results on header/body and header/sub-header paring under settings: (1) with all the font (full) and (2) removing top 50 popular body/sub-header fonts (non-popular) as described in Table \ref{Table:Tab07}. Classification thresholds are set by cross-validation with training data in each method respectively. 

\begin{table}[!t] \small
\centering \caption{Performance on binary classification of header/body and header/sub-header pairing under two settings as Table \ref{Table:Tab07}.}
\label{Table:Tab01} {
 \renewcommand{\arraystretch}{1.1}
     \begin{tabular}{c|cc|cc}
     \Xhline{1pt}
 task & \multicolumn{2}{c|}{header/body} &  \multicolumn{2}{c}{header/sub-header}  \\
 \hline
 setting& full&  non-popular & full& non-popular \\

   \Xhline{1pt}

Popularity &\underline{73.60}&55.29&68.04&61.35 \\



S-$k$NN &52.87&60.43&62.32&67.82\\


ConSim \cite{consim} & 55.81 &67.28&61.32&65.79  \\

SML \cite{cao2013similarity}&60.80 &\underline{67.34}&67.61&\underline{72.69}\\

DS-$k$NN &\textbf{76.93}&59.28&\textbf{71.30}&63.46\\

ASML &64.97 &\textbf{68.23} &\underline{68.43}&\textbf{73.41}\\

  \Xhline{1pt}
\end{tabular}
}
\end{table}

In setting ``full'', DS-$k$NN achieves the highest performance in both header/body and header/sub-header pairing. In header/body, Popularity achieves the second highest performance. It is consistent with the phenomenon shown in Figure \ref{fig:wikidis} that popular body fonts take a large proportion of head/body pairs in PDF designs. The main reason is that there are dominant popular fonts in ``full'' setting. Thus, in DS-$k$NN, the popular fonts will appear frequently in the candidate body set, and have more chances to hit the ground truth. In header/sub-header, ASML achieves the second highest performance.

To decrease the effects of dominant popular fonts, we also conduct the experiment under ``non-popular'' setting. In  setting ``non-popular'', ASML achieves the highest performance, followed by SML. Tables \ref{Table:Tab01} demonstrates the effectiveness of DS-$k$NN and ASML in both regular and non-popular font pairing tasks.


{An interesting observation is that DS-$k$NN performs much better in binary classification than in top-$N$ recommendation. In top-$N$ recommendation, according to the evaluation metric, the top recommended fonts have higher weights than the bottom ones. It means that if the top recommended fonts are not the same as ground truth, it is hard to achieve a high top-$N$ accuracy. In DS-$k$NN, since we rank all the test bodies with the similarity score compared with candidate bodies fonts, it has a high probability that fonts similar to the ground truth are ranked higher than the ground truth itself. It degrades the top-$N$ precision and recall score of DS-$k$NN. In binary classification, however, the performance would not degrade due to the order of recommendation.



\subsection{Subjective Evaluation}

Besides the quantitative evaluations, we also conduct subjective evaluation through user study on AMT and Upwork, which are crowdsourcing platforms targeting average people and professionals respectively. 

The study comprises a set of paired comparisons. One is either from DS-$k$NN or ASML, the other is from one of the compared methods. The users are only shown the sub-page contains the text as shown in Figure \ref{fig:visualization1}. We evaluate 500 comparisons and each comparison receives at least 11 ratings by average users and 3 ratings by designers.

Before describing the evaluation results, we firstly analyze the consistency of users' rating. If the users have consistent opinions about which pair is superior than the other, the ratings are more convincing and could be applied to the following studies.  It is important to analyze the rating consistency. If the users' ratings of two pairs are divergent on most of the comparisons, it shows that users do not have consistent opinions on font pairing task. Very likely the font pairing task is too subjective and could not be learnable. On the contrary, if the users have consistent opinions about which pair is superior than the other, the ratings are more convincing and could be applied to the following studies.

As described, we evaluate about 500 comparisons on AMT and Upwork and each comparison receives at least 11 ratings by average users and 3 ratings by designers. There are almost 150 average users and 10 designers in total. 

Suppose that there are $N$ comparisons, and for the $i$-th comparison, we denote the hits of pair1 as $hit1_i$ and the hits of pair2 as $hit2_i$. The normalized difference $d_i$ of the $i$-th comparison is as:

\begin{equation}
d_i = \frac {\lvert hit1_i - hit2_i \rvert } {hit1_i + hit2_i}.
\end{equation}

For example, assuming there are 11 ratings for one comparison, if the ratio of hits of two methods are 5:6, $d = (6-5)/(6+5) = 1/11 \simeq 0.09$. The value of $d$ is between 0 to 1. The higher the normalized difference, the higher the consistency is. 


To justify the users' ratings are consistent, we compare the distribution of users' rating with the distribution of pure random, and use hypothesis testing to test whether the two distributions are significantly distinct. 

We firstly introduce the rating consistency for average user on AMT. There are three steps. In the first step, we turn the continuous comparison results into binned data by grouping the comparisons into specified ranges according to $d$. We evenly divide [0,1] into six ranges from lowest to highest. The pdf of the normalized difference of users' ratings is shown Figure \ref{fig:pdf2} (a). 


\begin{figure}[t] 
            \centering
                \includegraphics[width=0.98\linewidth]{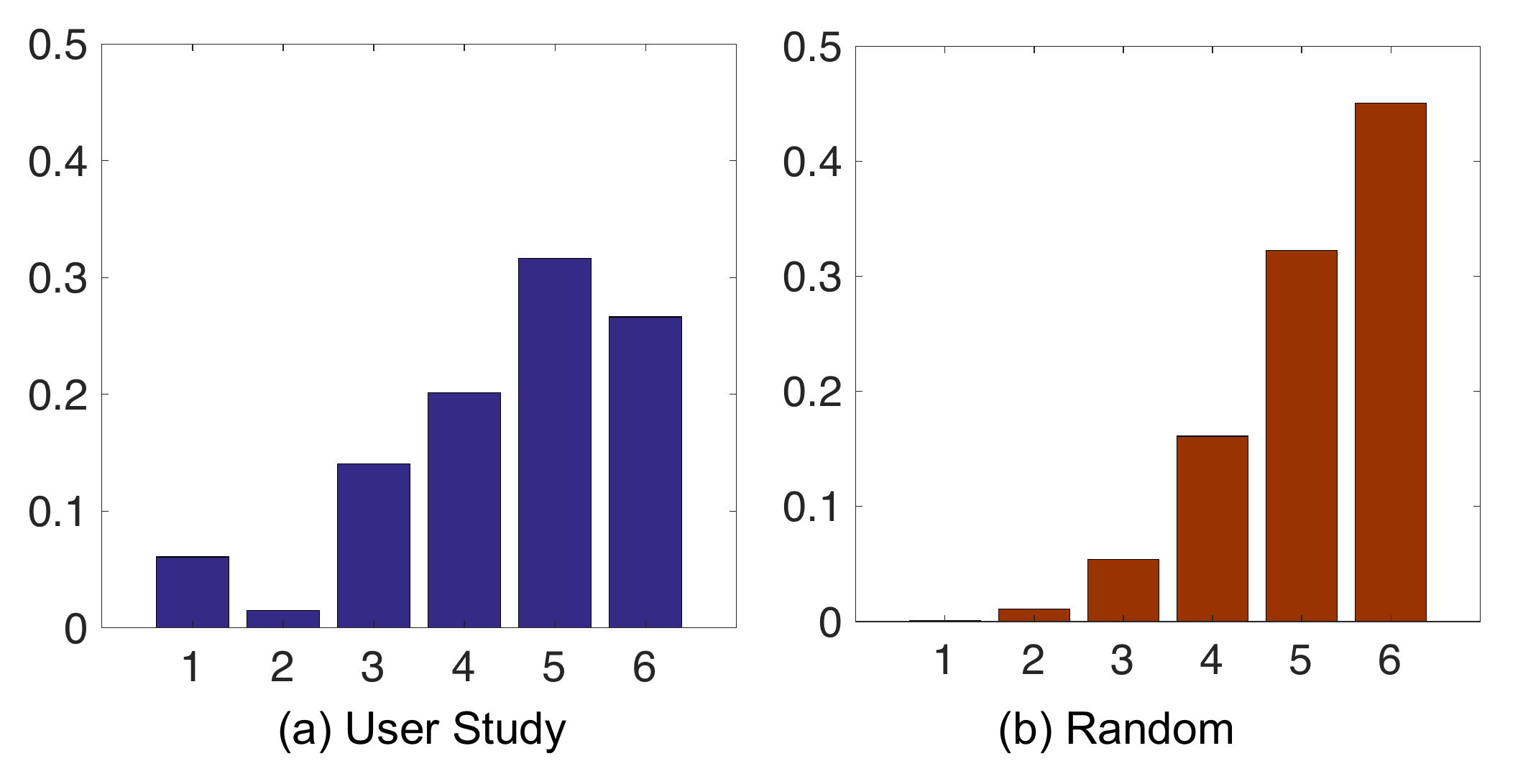}
            \caption{pdf of the normalized difference of average users' ratings (a) and pdf of pure random ratings (b). There are six bins for both pdfs. The x-axis from left to right demonstrates the consistency from highest to lowest. }
            \label{fig:pdf2}
\end{figure}

In the second step, we calculate the pdf of pure random choice analytically, shown in Figure \ref{fig:pdf2} (b).


In the third step, we apply $\chi^2$ hypothesis testing to test whether these two distributions are significantly distinct. Suppose that $n_j$ is the number of events observed in the $j$th bin, and that $e_j$ is the number expected according to random distribution. The $\chi^2$ statistic is calculated as:

\begin{equation}
\chi^2 =  \sum_{j} \frac { (n_j -e_j )^2 } {e_j}.
\end{equation}


Any term $j$ with $e_j = 0$ should be omitted from the sum. The average $\chi^2$ is 717.43 when we sum $j = 1$ to $6$. For $\chi^2$ testing, it is also suggested to omit the bins in which $e_j < 5$. In most cases, $e_1$ is very small in random distribution. Thus, we also calculate $\chi^2$ regarding $j = 2$ to $6$. The average $\chi^2$ is 117.32. According to $\chi^2$ distribution table, $\chi^2_{.005}$ = 16.750 under 6 bins and $\chi^2_{.005}$ = 14.860 under 5 bins \footnote{http://sites.stat.psu.edu/~mga/401/tables/Chi-square-table.pdf}. Thus we could safely draw the conclusion that the users' ratings are consistent and significantly different ($>99.95\%$) with pure random distribution.

We also analyze designer's rating consistency in the same way. In about 43$\%$ pairs, all the designers make the same choice (highest consistency). When calculating the pdf of pure random choices, only 25$\%$ pairs are with the highest consistency. In hypothesis testing, when comparing the pdf of designers' ratings and pure random ratings, we could safely make conclusion that the designers' choices are consistent and significantly different ($>99.95\%$) from pure random distribution.

\subsubsection{User Study Results}

\begin{figure}[t] 
            \centering
                \includegraphics[width=0.98\linewidth]{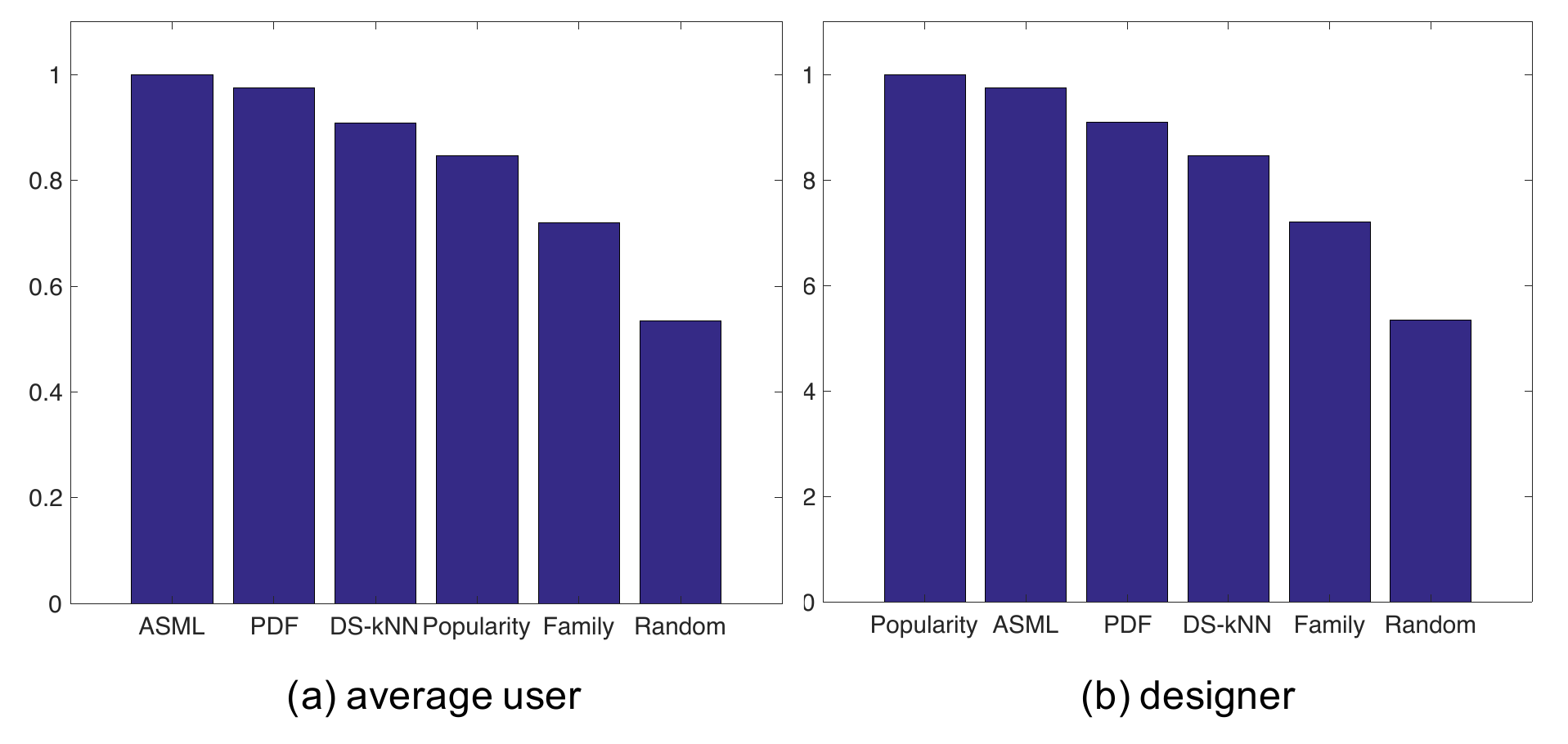}
            \caption{Subjective evaluation score of six methods by average user in (a) and designer in (b) by Bradley-Terry method.}
            \label{fig:userstudy_score}
\end{figure}


 We apply the Bradley-Terry models \footnote{http://sites.stat.psu.edu/~dhunter/code/btmatlab/} to get rankings for pairwise comparisons of ASML and DS-$k$NN to PDF, Random, Popularity and Family. The ranking scores of each methods based on average users' and designers' ratings are shown in Figure \ref{fig:userstudy_score}. For average user, the ranking results of these methods are ASML, PDF, DS-kNN, Popularity, Family and Random. For designer, the ranking results are Popularity, ASML, PDF, DS-kNN, Family, Random.

 Average user's ratings demonstrate that ASML outperforms hand-craft methods or even the pairs extracted from PDFs. Designers would prefer popular fonts most. We analyze that designers are more familiar with these popular fonts. However, as discussed before, only recommending popular fonts maybe less interesting to designers. Other ranks of designers are the same as average users'.  

}

\subsection{Users' Rating Prediction}


In this section, we want to evaluate the performance of predicting users' preference between pair1 (header A/sub-header B), and pair2 (header A/sub-header C).

\subsubsection{Experimental settings}
 

For each comparison, the ground-truth label is the pairing which receives a higher rating from user study. We predict users' choices by each method and compare the results with ground-truth labels as prediction accuracy of each method. For both average user and designer, we only use the ratings with the highest rating consistency as the evaluation set, which are more convincing.

We compare the performance of Popularity, S-$k$NN, ConSim, SML, DS-$k$NN and ASML. In Popularity, we compare the popularity of two sub-headers, and choose the more popular sub-header as the result. In S-$k$NN, we calculate the distance between header and each sub-header. We choose the pair with smaller distance as the result for S-$k$NN. The performance of ConSim, DS-$k$NN, SML, ASML are evaluated in a similar way as S-$k$NN, but with different scoring functions for calculating the distance between header and sub-header fonts. 


\subsubsection{Rating prediction results}

Table \ref{Table:Tab03} shows the accuracy of average users' and designers' ratings prediction with comparison methods under the highest consistency level. 


 For predicting average users' ratings, DS-$k$NN and ASML achieve the highest and second highest performance respectively. For predicting designers' ratings, DS-$k$NN, Popularity and ASML achieve the top 3 highest performance. It is generally consistent with the user study results in Section 5.5.

When looking into S-$k$NN under average users' and designers' ratings, it is interesting to see that average users prefer the pairing with similar header and sub-header fonts, while designers prefer the pairing with contrast header and sub-header fonts. It shows the hardness of predicting both tasks in the uniform scoring function. However, ASML could achieve the second and the third highest performance in both tasks, which shows the effectiveness of the learned scoring function in ASML.



\begin{table}[!t] \small
\centering \caption{Accuracy of predicting average users' and designers' ratings with comparison methods.}
\label{Table:Tab03} {
 \renewcommand{\arraystretch}{1.1}
    \begin{tabular}{l*{5}{p{2cm}<{\centering}}}
     \Xhline{1pt}

& average user &designer\\

   \Xhline{1pt}  

Popularity &55.56 & \underline{57.89}\\



S-$k$NN &55.33&45.45\\

ConSim & 54.32&  52.15 \\


SML &56.22&54.59\\

DS-$k$NN(ours1) &\textbf{68.18}&\textbf{59.81}\\

ASML(ours2)&\underline{58.67} &56.94\\

  \Xhline{1pt}
\end{tabular}
}
\end{table}


\section{Conclusion}
In this paper, we introduced the problem of visual font pairing. To our best knowledge, it is the first time automatic font pairing has been addressed in multimedia and computer vision field. We introduced a new database called FontPairing, from millions of PDF documents on the Internet. We automatically extracted header/sub-header, header/body pairs from PDF pages. We proposed two automatic font pairing methods through learning fine-grain visual relationships from large-scale human-generated font pairs: dual-space $k$-NN and asymmetric similarity metric learning. Comparisons are conducted against several baseline methods based on rules from professional designers. Experiments and user studies demonstrate the effectiveness of our proposed dataset and methods.

\bibliographystyle{IEEEtran}
\bibliography{fontpairing}

\begin{IEEEbiography}[{\includegraphics[width=1in,height=1.25in,keepaspectratio]{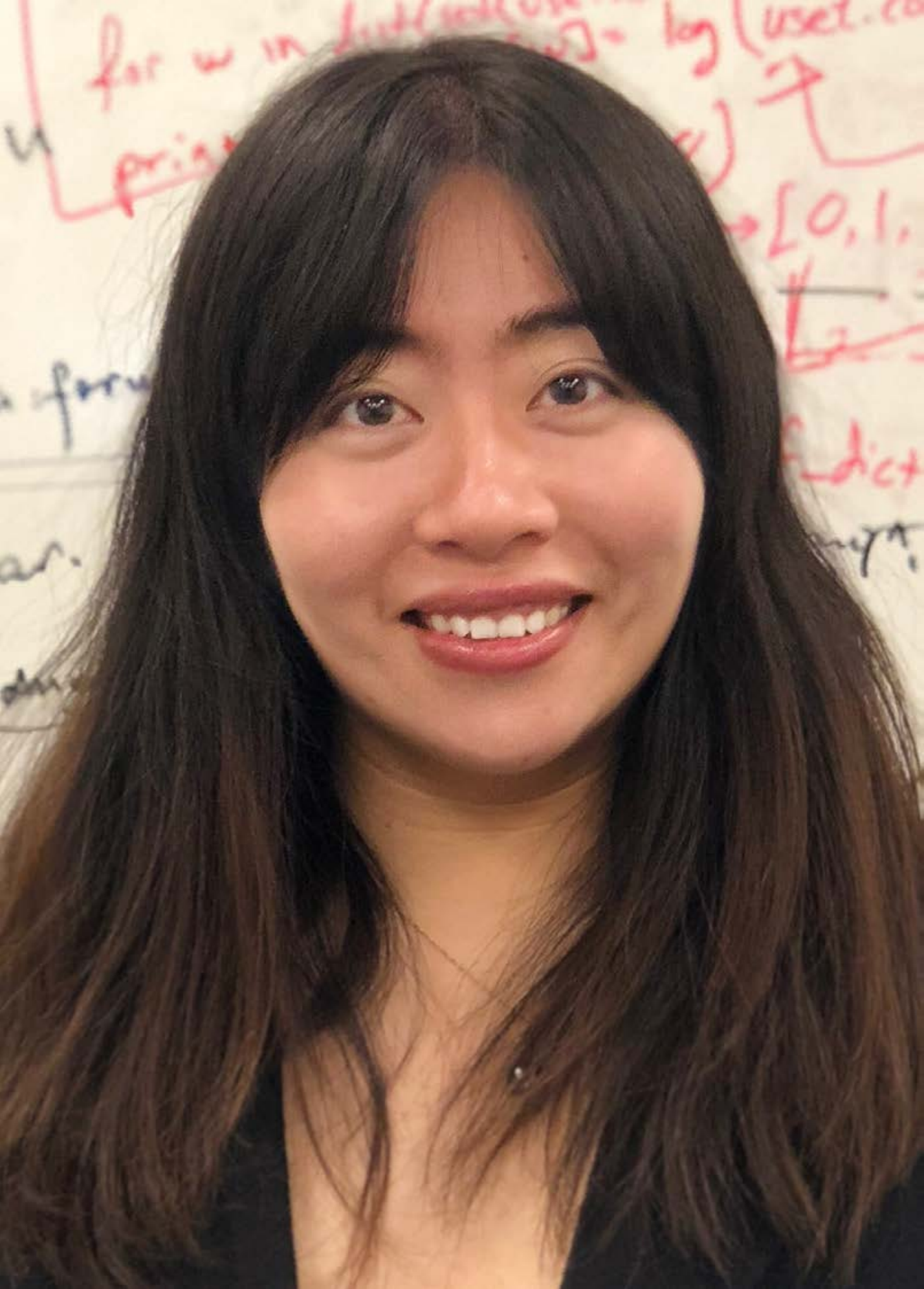}}]{Shuhui Jiang} received the B.S. and M.S. degrees in Xi'an Jiaotong University, Xi'an, China, in 2007 and 2011, respectively, and the Ph.D. degree in electrical and computer engineering from School of Electrical and Computer Engineering, Northeastern University (Boston, USA). She was the recipient of the Dean's Fellowship of Northeastern University from 2014. She is interested in machine learning, multimedia and computer vision. She has served as the reviewers for IEEE journals: IEEE Transactions on Multimedia, IEEE Transactions on Neural Networks and Learning Systems, etc. She was a research intern with Adobe research lab, San Jose, US, in summer 2016.
\end{IEEEbiography}

\begin{IEEEbiography}[{\includegraphics[width=1in,height=1.25in,keepaspectratio]{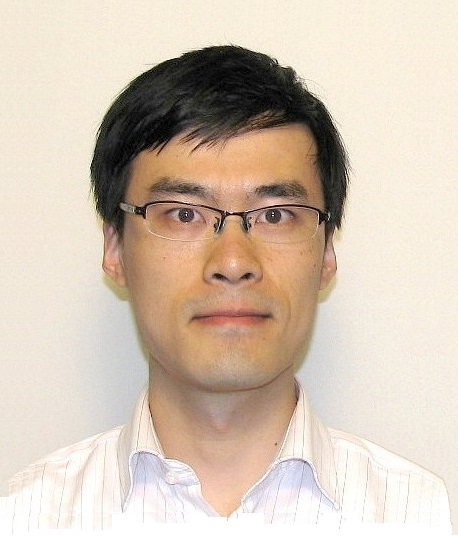}}]{Zhaowen Wang} (M'14) received the B.E. and M.S. degrees  from  Shanghai  Jiao  Tong  University,  China,  in  2006  and  2009,  respectively,  and  the  Ph.D. degree in electrical and computer engineering from the  University  of  Illinois  at  Urbana-Champaign,  in 2014. He is currently a Senior Research Scientist with the Creative Intelligence Lab,  Adobe Inc.  His research  has  been  focused  on understanding and enhancing images, videos and graphics via machine learning algorithms, with a particular interest in sparse coding and deep learning.
\end{IEEEbiography}

\begin{IEEEbiography}[{\includegraphics[width=1in,height=1.25in,keepaspectratio]{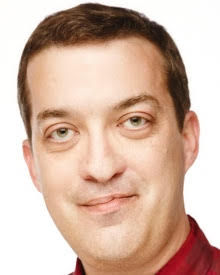}}]{Aaron Hertzmann} is a Principal Scientist at Adobe and an ACM Distinguished Scientist. He received a BA in Computer Science and Art/Art History from Rice University in 1996, and a PhD in Computer Science from New York University in 2001. He was a Professor at University of Toronto from 2003 to 2013, and has also worked at Pixar Animation Studios, University of Washington, Microsoft Research, Mitsubishi Electric Research Lab, Interval Research Corporation and NEC Research. He was an Associate Editor for \textit{ACM Transactions on Graphics}, for ten years. His awards include the MIT TR100 (2004), a Sloan Foundation Fellowship (2006), a Microsoft New Faculty Fellowship (2006), the CACS/AIC Outstanding Young CS Researcher Award (2010), and the Steacie Prize for Natural Sciences (2010), as well as several conference best paper awards.
\end{IEEEbiography}

\begin{IEEEbiography}[{\includegraphics[width=1in,height=1.25in,keepaspectratio]{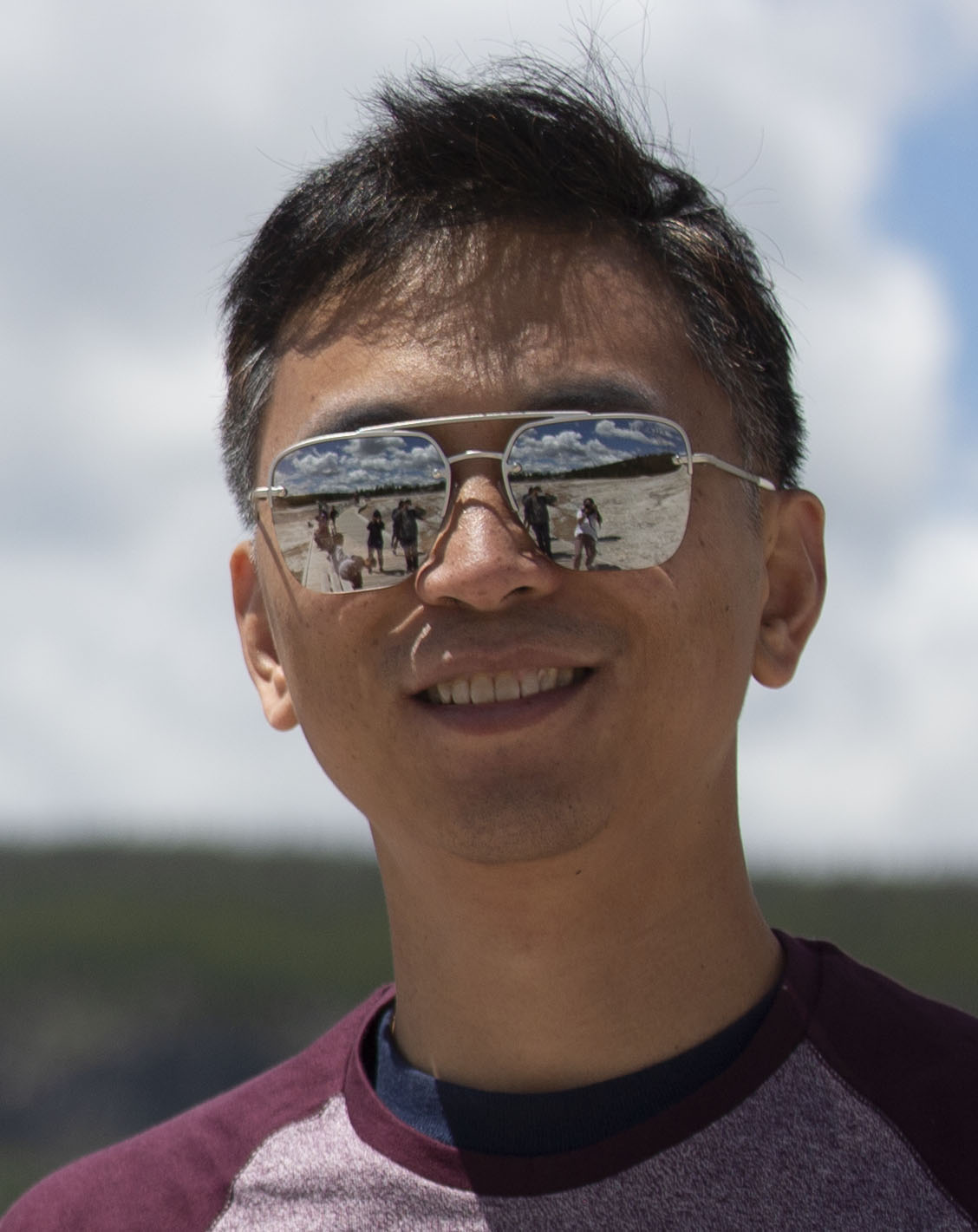}}]{Hailin Jin} received the bachelor’s degree in Automation from Tsinghua University, Beijing, China, in 1998, and the M.S. and D.Sc. degrees in Electrical Engineering from Washington University in Saint Louis, in 2000 and 2003, respectively. From fall 2003 to fall 2004, he was a Post-Doctoral Researcher at the Computer Science Department, University of California at Los Angeles. Since 2004, he has been with Adobe Research, where he is currently a Senior Principal Scientist. He received the Best Student Paper Award (with J. Andrews and C. Sequin) at the 2012 International CAD Conference for work on interactive inverse 3D modeling. He is a member of the IEEE Computer Society.
\end{IEEEbiography}

\begin{IEEEbiography}[{\includegraphics[width=1in,height=1.25in,clip,keepaspectratio]{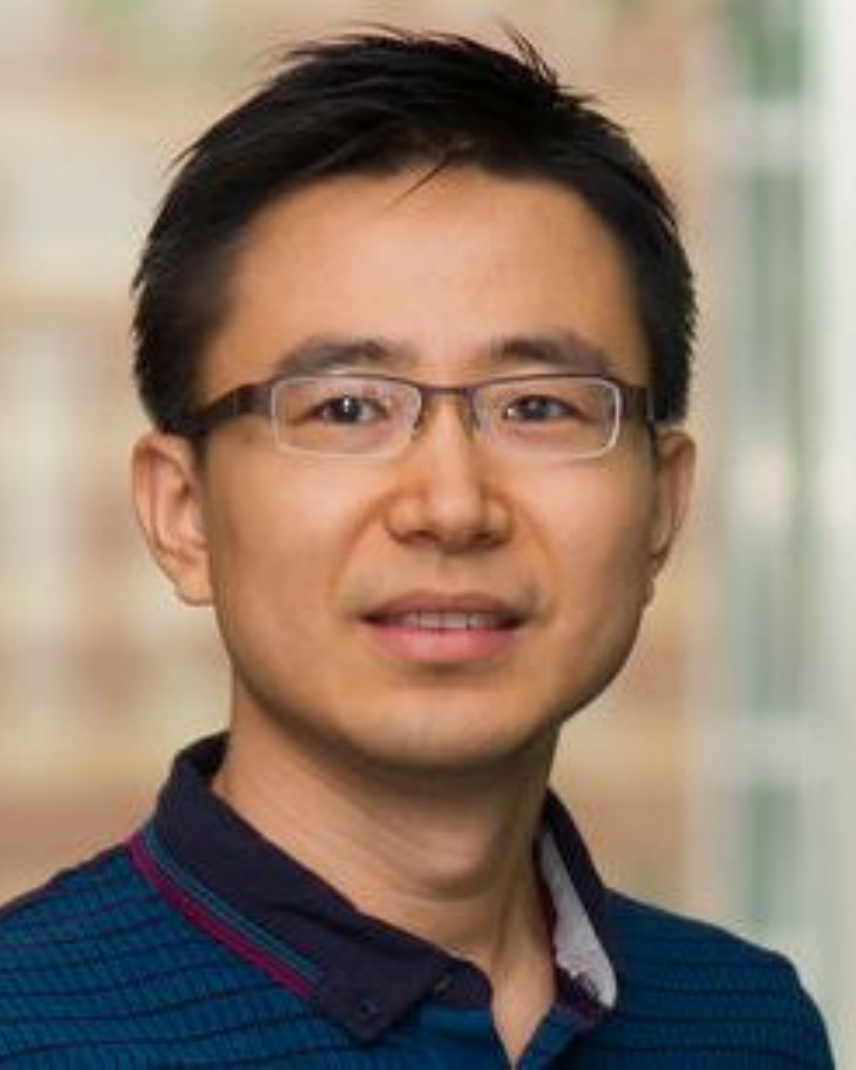}}]{Yun Fu} (S'07-M'08-SM'11) received the B.Eng. degree in information engineering and the M.Eng. degree in pattern recognition and intelligence systems from Xi'an Jiaotong University, China, respectively, and the M.S. degree in statistics and the Ph.D. degree in electrical and computer engineering from the University of Illinois at Urbana-Champaign, respectively. He is an interdisciplinary faculty member affiliated with College of Engineering and the College of Computer and Information Science at Northeastern University since 2012. His research interests are Machine Learning, Computational Intelligence, Big Data Mining, Computer Vision, Pattern Recognition, and Cyber-Physical Systems. He has extensive publications in leading journals, books/book chapters and international conferences/workshops. He serves as associate editor, chairs, PC member and reviewer of many top journals and international conferences/workshops. He received seven Prestigious Young Investigator Awards from NAE, ONR, ARO, IEEE, INNS, UIUC, Grainger Foundation; nine Best Paper Awards from IEEE, IAPR, SPIE, SIAM; many major Industrial Research Awards from Google, Samsung, and Adobe, etc. He is currently an Associate Editor of the IEEE Transactions on Neural Networks and Leaning Systems (TNNLS). He is fellow of IAPR, OSA and SPIE, a Lifetime Distinguished Member of ACM, Lifetime Member of AAAI and Institute of Mathematical Statistics, member of ACM Future of Computing Academy, Global Young Academy, AAAS, INNS and Beckman Graduate Fellow during 2007-2008.
\end{IEEEbiography}




\end{document}